# REAL: Rapid Exploration with Active Loop-Closing toward Large-Scale 3D Mapping using UAVs

Eungchang Mason Lee[1], *Student Member, IEEE*, Junho Choi[1],
Hyungtae Lim[1], *Student Member, IEEE*, and Hyun Myung[1], *Senior Member, IEEE*

*Abstract*— Exploring an unknown environment without colliding with obstacles is one of the essentials of autonomous vehicles to perform diverse missions such as structural inspections, rescues, deliveries, and so forth. Therefore, unmanned aerial vehicles (UAVs), which are fast, agile, and have high degrees of freedom, have been widely used. However, previous approaches have two limitations: a) First, they may not be appropriate for exploring large-scale environments because they mainly depend on random sampling-based path planning that causes unnecessary movements. b) Second, they assume the pose estimation is accurate enough, which is the most critical factor in obtaining an accurate map. In this paper, to explore and map unknown large-scale environments rapidly and accurately, we propose a novel exploration method that combines the pre-calculated *Peacock Trajectory* with graph-based global exploration and active loop-closing. Because the two-step trajectory that considers the kinodynamics of UAVs is used, obstacle avoidance is guaranteed in the receding-horizon manner. In addition, local exploration that considers the frontier and global exploration based on the graph maximizes the speed of exploration by minimizing unnecessary revisiting. In addition, by actively closing the loop based on the likelihood, pose estimation performance is improved. The proposed method's performance is verified by exploring 3D simulation environments in comparison with the state-of-the-art methods. Finally, the proposed approach is validated in a real-world experiment.

## I. INTRODUCTION

In recent years, autonomous robots have been utilized to inspect structures and explore hazardous areas or disaster sites [1], [2]. In particular, because of the agile movement and high degree of freedom, exploration in unknown environments using unmanned aerial vehicles (UAVs) has been widely researched [3]-[10]. However, existing exploration methods using UAVs mainly depend on random sampling-based path planning, such as Rapidly-exploring Random Trees (RRT) [11], which yields unnecessary movements that degrade the speed of exploration [8], [9]. Moreover, the trajectory to be executed hardly considers the kinodynamics of UAVs, which engenders the infinite torque input. In addition, because most of them ignore the drift of the pose estimation, such as through Visual-Inertial Odometry (VIO), inaccurate maps may be obtained, especially when exploring large-scale environments. Therefore, a rapid and efficient path

[1]Eungchang Mason Lee, [1]Junho Choi, [1]Hyungtae Lim, and [1]Hyun Myung are with School of Electrical Engineering, KI-AI, and KI-R at KAIST (Korea Advanced Institute of Science and Technology), Daejeon, 34141, Republic of Korea. {eungchang_mason, cjh6685kr, shapelim, hmyung}@kaist.ac.kr

This work has been supported by the Unmanned Swarm CPS Research Laboratory program of Defense Acquisition Program Administration and Agency for Defense Development.(UD190029ED) planner that considers the limited operating time (battery), the trajectory considering UAVs' kinodynamics, and the pose estimation must be regarded in exploration.

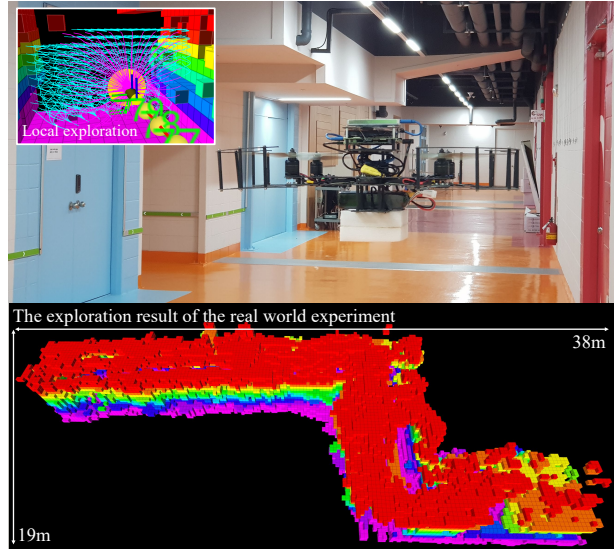

Fig. 1. An instant of the exploration in the real world. The UAV explores a confined area with a dimension of $38\times19\times3$ m$^3$ located under the KI Building from KAIST, Daejeon, South Korea.

To this end, in this paper, a novel exploration method called *Rapid Exploration with Active Loop-closing*, *REAL*, is proposed, combining Peacock Trajectory that deviates from the random sampling method with graph-based global exploration and active loop-closing. First, local exploration is executed depending on the scoring matrix regarding the obstacle avoidance and the frontier via the two-step trajectory considering the kinodynamics of the UAV. Second, graph-based global exploration is performed through a gain function that considers the frontier clusters. Third, actively closing the loop leveraging the likelihood-based selection is developed to reduce the drift of pose estimation, which is important for accurately mapping large-scale environments.

The performance of REAL is validated by comparing it with the state-of-the-art exploration methods experimentally in small- and large-scale simulation environments based on Gazebo [21]. Furthermore, a real-world experiment is conducted to demonstrate the capability of REAL.

The main contributions of this paper are fourfold:

- The control-efficient Peacock Trajectory, which considers the kinodynamics of UAVs and avoids inefficient random sampling, is used to proceed with the local exploration based on the scoring matrix considering obstacle avoidance and the frontier. It assumes maximum initial and final velocities, which increase the speed of exploration, and indirectly contributes to stable pose estimation thanks to its smoothness.

- A graph-based global exploration planner is developed utilizing the optimal A* algorithm [14] with the cost function considering the frontier clusters, only for collision, dead-end, or redundant revisiting situations. It prevents unnecessary back-and-forth movements of the pure frontier-based method and maximizes planning efficiency to increase exploration speed.
- Active loop-closing via maximizing the likelihood is proposed to enhance the performance of the pose estimation, which is especially essential for exploring large-scale environments. To the best of our knowledge, this is the first study to adopt active loop-closing within exploration for UAVs.
- Evaluation and demonstration of the proposed approach in both simulation and a real-world experiment are conducted.

The rest of this paper is structured as follows: Section II summarizes the related work. Section III defines the problem and describes the proposed approach in detail. Evaluation during simulation and validation in a real-world experiment are discussed in Section IV before we conclude the presented work and describe future works in Section V.

## II. RELATED WORK

To explore an unknown environment, a frontier-based exploration method [3] was proposed, which defined the frontier as the free area adjacent to the unknown area. However, this method explores the entire space by tracking the nearest frontier repeatedly, which leads to unnecessary back-and-forth movements [4], [9]. Cieslewski *et al.* [4] proposed a variant of the frontier-based method that prioritizes the frontier in a way that minimizes the change of velocity of UAVs to maintain high velocity. Although it is conceptually similar to our proposed method, it solely takes the velocity into account; thus, it does not directly address the kinodynamics of UAVs. However, our proposed exploration method is based on minimum snap trajectory [15], which helps UAVs move in an energy-efficient way.

To tackle the potential limits of the frontier-based method, some researchers have conducted random sampling-based methods. Next-best-view planner (NBVP) [5] is a renowned method, which repeats moving only the first edge of the RRT with the highest information gain in the free region. In addition, Dang *et al.* [6] proposed a graph-based path planning (GBP) that leverages Rapidly-exploring Random Graph (RRG) to move to the position with the highest gain in the local subspace for navigating in subterranean environments. In addition, global exploration was considered for dead-end situations. In motion primitives-based path planning (MBP) [7], after generating motion primitives to the randomly sampled configuration state, the UAV moves to a place with the highest gain. It guarantees obstacle avoidance by selecting a future safe trajectory. Although it adopts motion primitives as trajectories, their systems are assumed to be discrete systems; thus, acceleration input is fixed during the unit interval. The method we propose is different in that continuous input is considered. Many suboptimal and unnecessary movements are found in NBVP, GBP, and MBP because they strongly depend on random sampling [8], [9].

Exploration methods that leverage the advantages of random sampling and the frontier have also been studied. Witting *et al.* [8] proposed a method that memorizes the visited place from NBVP [5]. In a dead-end situation, RRT is built using nodes in history and moves to a position that considers the frontier to improve the exploration speed. Efficient autonomous exploration planning (AEP) [9] fused NBVP for local exploration and the frontier for global exploration to increase the efficiency of spatial visiting.

However, the aforementioned methods [4]-[9] are based on the premise that pose estimation is accurate enough, so it does not affect the performance of exploration. To consider the localization, Papachristos *et al.* [10] proposed a method that divides NBVP into two stages; selecting the region with the highest gain, and then selecting a position with the minimum uncertainty of localization again in the nearby area. Stachniss *et al.* [12] and Lehner *et al.* [13] proposed exploration methods that consider active loop-closing to correct the drift of estimated pose for ground vehicles.

In summary, we propose a novel, rapid, frontier-based exploration method with minimum snap trajectory, solving the limitations of random sampling-based methods, which is followed by active loop-closing to minimize drift to achieve successful localization.

## III. PROPOSED APPROACH

The problem to be addressed in this paper can be summarized as follows: The three-dimensional position vector $\vec{p}$ is laid in a bounded volume $\mathcal{V}$; that is, $\vec{p} \in \mathcal{V} \subset \mathbb{R}^3$. The bounded volume $\mathcal{V}$ is classified into unknown $\mathcal{V}_{\text{un}}$, occupied $\mathcal{V}_{\text{occ}}$, and free $\mathcal{V}_{\text{free}}$. Initially, the entire volume is set as $\mathcal{V} = \mathcal{V}_{\text{un}}$. Afterward, the ideal condition after exploration is defined as $\mathcal{V} = \mathcal{V}_{\text{occ}} \cup \mathcal{V}_{\text{free}}, \mathcal{V}_{\text{un}} = \varnothing$, while the volume is mapped by collecting sensor data from the estimated pose. Therefore, our goal is to a) remove $\mathcal{V}_{\text{un}}$ as much as possible and discern $\mathcal{V}_{\text{free}}$ and $\mathcal{V}_{\text{occ}}$ at the same time, b) decrease exploration time in consideration with the limited operating time, and c) guarantee accurate pose estimation.

*A. Algorithm Overview*

Our proposed method, REAL, mainly consists of three parts: a) a local exploration planner, b) a global exploration planner, and c) an active loop-closing, as illustrated in Fig. 2. First, Peacock Trajectory is calculated once at the beginning of the algorithm, which calculates a scoring matrix that allows local exploration to proceed while the graph is built with the nodes. The graph-based global exploration is conducted to prevent redundant revisiting through a gain function. Additionally, the likelihood of the loop is calculated to close the loop actively.

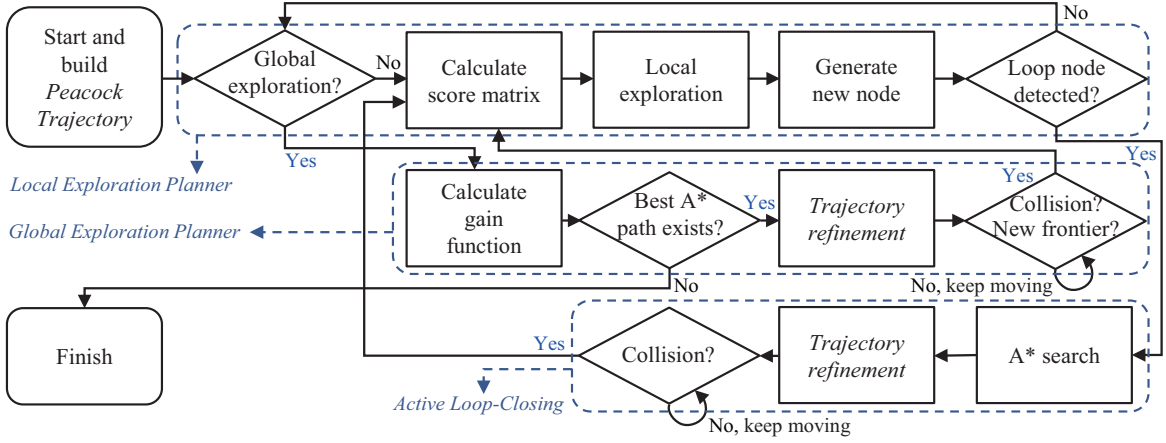

Fig. 2. The flowchart of the proposed approach, REAL. The boxes with dashed lines represent *Local Exploration Planner*, *Global Exploration Planner*, and *Active Loop-Closing* blocks, respectively. The algorithm starts by building *Peacock Trajectory* to calculate the score matrix to proceed with local exploration. The graph with the nodes is updated at regular intervals to find the node with the maximum likelihood to close the loop actively. In addition, the graph is used to find the best global path based on the gain function considering the frontier clusters.

### B. 3D Frontier

In this work, OctoMap [20] is used to manage the volume efficiently with the centroids of cells. The frontier is defined as a set of free cells adjacent to at least one unknown cell [3], and the adjacent area means that the Euclidean distance between the centroids is equal to the resolution of OctoMap, $\varrho_{oct}$. The frontier $\mathcal{F}$ is defined as follows:

$$\mathcal{F} = \{\vec{p}_f \mid \vec{p}_f \in \mathcal{V}_{\text{free}}, \ \exists \vec{p}_u \in \mathcal{V}_{\text{un}}, \ \|\vec{p}_f - \vec{p}_u\| = \varrho_{oct}\}. \quad (1)$$

The frontier can be separated into $\mathcal{F}_{\text{new}}$, which is newly generated at every depth sensor frame (an RGB-D camera in this work) and $\mathcal{F}_{\text{map}}$, which is continuously stored in the map. $\mathcal{F}_{\text{new}}$ and $\mathcal{F}_{\text{map}}$ are used in local and global exploration, respectively.

### C. Graph Representation

During local exploration, the current state is encoded into the $K$-th node, $n_K$, which is represented as the four-dimensional state $(\vec{p}_K, \psi_K)$, where $\psi_K$ denotes the yaw angle, at regular intervals. Let $k$ be the index of a node $n_k$, then all previous nodes and the current node are expressed as graph representation as $\mathcal{N}_K = \{n_1, \cdots, n_k, \cdots, n_K\}$. Simultaneously, edges are connected if two conditions are met: a) the path between the nodes is free and b) the distance is closer than the maximum sensor range, $r_{\max}$, while storing the length of the edges, $\ell$. The graph is used in the global exploration planner, adopting $\ell$ as the cost for the path. Visual descriptions of the graph can be seen in Fig. 3(a). In addition, the likelihood for the loop is calculated at every node generation (see Section III.G).

### D. Peacock Trajectory

To consider the kinodynamics of the UAVs and deviate from the random sampling-based path planning, which yields inefficient movements, *Peacock Trajectory* is employed in two steps with control-efficient minimum snap trajectories [15]. It was developed in our early work [19] solely to check for collisions and unknown regions; its name derives from its appearance. Note that the main difference with our early work is that ours involves the newly generated frontier $\mathcal{F}_{\text{new}}$. Because the input of UAVs is algebraically related to the fourth derivative of position, the snap [15], tracking the minimum snap trajectory entails efficient control, possibly extending battery life by avoiding infinite torque from discontinuities in the snap. The minimum snap trajectory can be calculated by minimizing the objective function as follows:

$$x^*(t) = \operatorname*{argmin}_{x(t)} \int_{t_0}^{t_0+T} \mathcal{L} \, dt, \quad \mathcal{L} = (x^{(4)})^2 = s^2 \quad (2)$$

where $x(t)$ and $x^*(t)$ denote the position and snap-optimized position over time, respectively. $\mathcal{L}$ is the objective function that should be minimized over time $T$, and $s$ is the snap. This can be optimized by solving the Euler-Lagrange equation as follows:

$$\frac{\partial \mathcal{L}}{\partial x} - \frac{d}{dt}\left(\frac{\partial \mathcal{L}}{\partial \dot{x}}\right) + \frac{d^2}{dt^2}\left(\frac{\partial \mathcal{L}}{\partial \ddot{x}}\right) + \cdots \\ + (-1)^n \frac{d^n}{dt^n}\left(\frac{\partial \mathcal{L}}{\partial x^{(n)}}\right) = 0 \quad (3)$$

where $n = 4$ due to the property of the snap. As a result, $x^*(t)$ can be obtained as the 7th order polynomial function of time. By setting the conditions with the desired position, velocity, acceleration, and jerk at the initial and final time of the trajectory, the coefficients of the polynomial can be easily found. Part of the conditions can be set to consider the sensor's Field of View (FoV) as follows:

$$\begin{cases} x(T)_{(i,j)} = x(t_0) + v_{\max}T\cos(\psi_j)\cos(\theta_i) \\ x(2T)_{(i,j,b)} = x(T)_{(i,j)} + v_{\max}T\cos(\bar{\psi}_b) \end{cases} \quad (4)$$

where $\psi_j$ and $\bar{\psi}_b$ denote the yaw angles for the first step and the second step trajectories, respectively, and $\theta_i$ represents the pitch angle of the first step. In this way, the coefficients of the polynomials for the $x$-, $y$-, and $z$-axes can be computed separately. Note that the same initial and final velocities as maximum, $v_{\max}$, are assumed for the fast navigation.

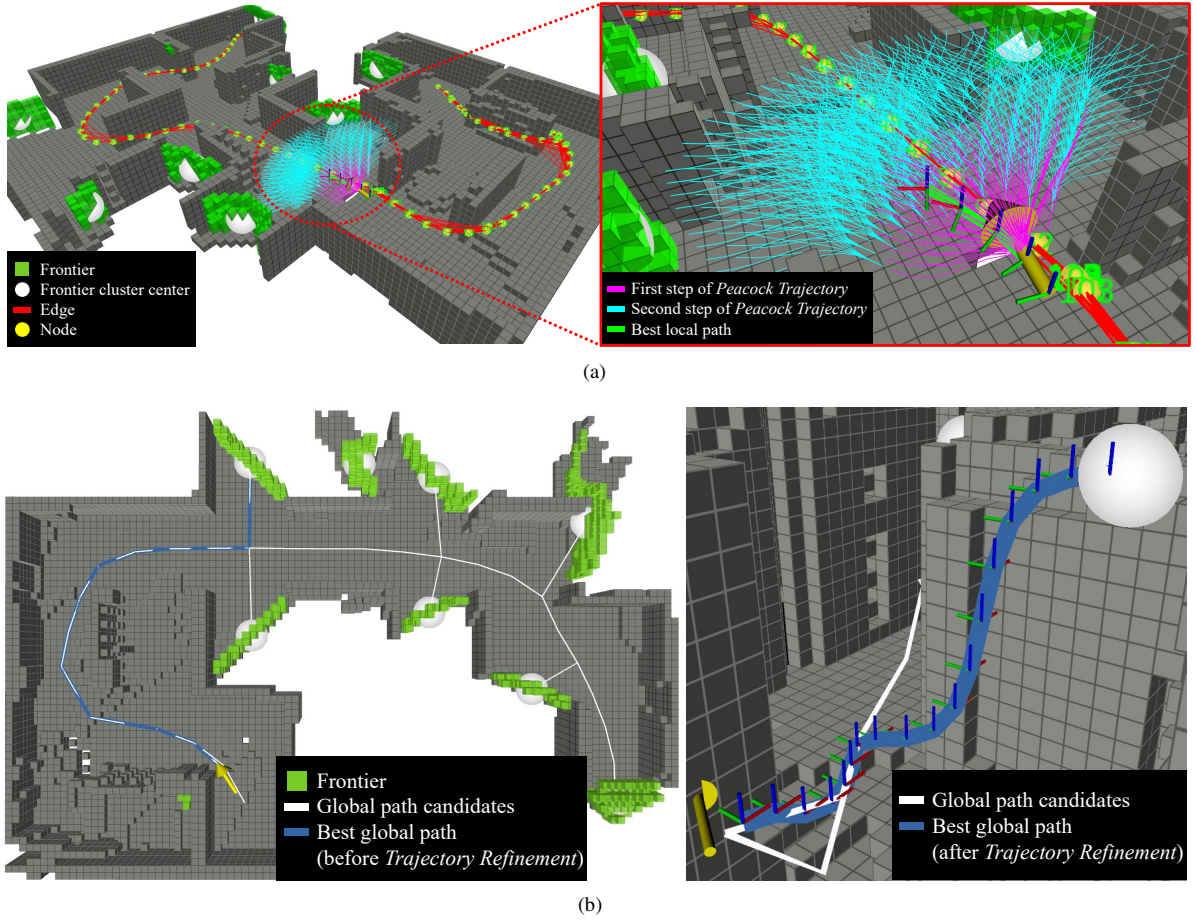

Fig. 3. Visual descriptions of (a) local exploration and (b) global exploration in a small-scale simulation map. Peacock Trajectory is colored in turquoise and magenta and the thick, green line denotes the best local path in (a). The global exploration that considers the frontier clusters is shown in (b), with the best global path shown as a thick, blue line. The smooth trajectory after Trajectory Refinement also can be seen in the right part of (b).

The interval, $T$, can be set with the length of trajectory $\ell_{\text{traj}}$, or $T = \frac{\ell_{\text{traj}}}{v_{\text{max}}}$. Consequently, four-dimensional Peacock Trajectory including $(\vec{p}(t), \psi(t))$ is generated, where $\psi(t)$ is computed as forward direction from the velocities.

Although motion primitives [7], [22], [23] are similar in appearance, they assume the fixed input over the unit interval, which means the discrete system. In contrast, Peacock Trajectory is based on the continuous system that continuously considers different input from the optimized snap and consists of two steps inspired by the receding-horizon manner to avoid unknown suboptimal situations (e.g., dead-end or sudden obstacles outside of sensor range) from directly moving to the end of entire trajectories. Likewise, although path groups used in [24] are similar in appearance, each path is generated as a cubic spline curve that does not consider the kinodynamics of UAVs.

One additional benefit of utilizing Peacock Trajectory is that a smooth trajectory indirectly increases the VIO performance and hence the mapping quality by properly exciting IMU and providing continuous feature tracking, which was analyzed in the our previous work [18]. To use it, the precomputation of the coefficients is conducted only once at the beginning of the proposed algorithm, which takes around 2 ms using the general onboard computers. The visualization of it can be seen in Fig. 3(a).

*E. Local Exploration Planner*

Thereafter, the best local trajectory among Peacock Trajectory is chosen in terms of collision avoidance and the number of frontiers. To do so, two matrices, $\Lambda$ and $\Gamma$, with the same dimension as the first-step trajectory, are introduced. The collision check matrix $\Lambda$ and frontier-considering matrix $\Gamma$ are initialized to zero and one, respectively. Both the first- and the second-step trajectories are used to check collision. If the first-step trajectory is laid on the $\mathcal{V}_{\text{free}}$, the second-step trajectories are checked to see whether they are free to accept score $\lambda$. That is, $\Lambda_{(i,j)} \mathrel{+}= \lambda$. Otherwise, $\Lambda_{(i,j)} = 0$.

For every $\vec{p}_f \in \mathcal{F}_{\text{new}}$, the location of the nearest first-step trajectory's end point is found to add score $\gamma$. That is, $\Gamma_{(i_\gamma, j_\gamma)} \mathrel{+}= \gamma$, where $(i_\gamma, j_\gamma) = \underset{(i,j)}{\operatorname{argmin}} \|\vec{p}_f - \vec{p}(T)_{(i,j)}\|$. Here, the second-step trajectories are not used because only the first-step trajectory is executed in the receding-horizon manner. Then, the scoring matrix $\mathbb{G}_L$ is defined as the Hadamard product between $\Lambda$ and $\Gamma$ as follows:

$$\mathbb{G}_L = \Lambda \odot \Gamma, \quad \odot : \text{Hadamard product}. \tag{5}$$

The highest $\mathbb{G}_L$ location is found to track the first step of Peacock Trajectory, which is the best local path that is shown in Fig. 3(a) as a thick, green line. This simple scoring policy only needs $O(\log n)$ computational complexity for octree search to guarantee complete obstacle avoidance. In addition, by tuning $\gamma$ and $\lambda$, the priority between the frontier and collision can be set.

Afterward, the global exploration proceeds if one of two conditions are met: a) if $\mathcal{F}_{\text{new}} = \emptyset$ and no $\mathcal{F}_{\text{map}}$ is visible inside of the sensor FoV (i.e., dead-end or redundant revisiting) or b) if $\mathcal{F}_{\text{new}} = \emptyset$ and $\max(\mathbb{G}_L) = 0$ (collision).

### F. Global Exploration Planner

To find the best global path $\xi_G$ in consideration with $\mathcal{F}_{\text{map}}$, $\mathcal{F}_{\text{map}}$ is clustered into a set $\mathbb{C} = \{\mathcal{C}_1, \cdots, \mathcal{C}_w, \cdots, \mathcal{C}_W\}$ with the center point $\vec{p}_w$ and the size $S_w$ of each of the $W$ clusters. Then, $n_w \in \mathcal{N}_K$ and $\mathcal{C}_w$ with the closest free path among possible combinations are assigned in pairs. With the pairs, the A* algorithm is executed to find the set of shortest paths $\Xi$ from $n_K$ to $n_w$. Finally, the best global path $\xi_G$ is determined based on the gain function as follows:

$$\xi_G = \underset{\xi_w \in \Xi}{\operatorname{argmax}} \mathbb{G}_G(\xi_w), \quad \mathbb{G}_G(\xi) = S e^{-\zeta \ell_{tot}} \quad (6)$$

where $\zeta$ represents a hyperparameter to balance the weight between the total distance of the path $\ell_{tot}$ and the corresponding frontier cluster size. An instance of global exploration planning is shown in Fig. 3(b), where straight, white lines represent $\xi_w \in \Xi$. The centroids $\vec{p}_w$ of $\mathcal{C}_w \in \mathbb{C}$ are visualized as white spheres. It can be checked that the paths are obtained only when free.

The global exploration is immediately stopped to proceed with the local exploration again when $\mathcal{F}_{\text{new}}$ is generated within sensor FoV or a possible collision is detected through Peacock Trajectory, increasing the exploration speed by maximizing the efficiency of visiting. Similarly, it prevents unnecessary back-and-forth movements generated from pure frontier-based methods, because it is conducted only for dead-end or redundant revisiting and collision situations. Because $\xi_G$ is merely the set of edges that are straight lines and does not consider kinodynamics of UAVs, it is refined into the minimum snap trajectory before the move, which will be detailed in the later subsection.

### G. Active Loop-Closing

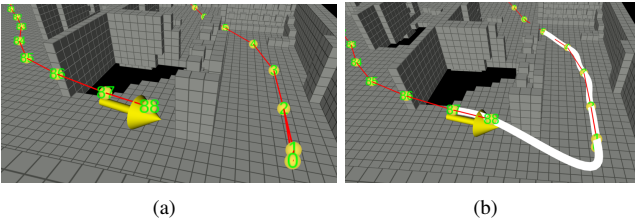

Fig. 4. An instance of active loop-closing. The red lines are edges between nodes that are represented as numbered yellow spheres. The thick, white line is the loop-closing path $\xi_{LC}$ after Trajectory Refinement. The node $n_0$ is selected as $n_{LC}^*$ from the current node $n_{88}$.

As previously mentioned, loop-closing is decided during the node generation. When generating $n_K$, if the geodesic distance between $n_K$ and $n_k' \in \mathcal{N}_{K-1}$ is far enough but physically closer than $r_{\max}$, then the likelihood for the loop is calculated to select and move toward the optimal node. To find the optimal node $n_{LC}^*$, which is with the highest probability of the loop, one possible implementation of the likelihood considering the geodesic distance between nodes is defined as follows:

$$\begin{aligned} L(\Theta; \mathcal{X}) &= f(\mathcal{X}|\Theta) \\ &\equiv \mathcal{H}(|K - k'| - \tau(v_{\max}))e^{-|\psi_{n_K} - \psi_{n_{k'}}|} \end{aligned} \quad (7)$$

where $\Theta$ and $\mathcal{X}$ respectively represent all parameters of the proposed system and a random variable that denotes how likely corresponding two nodes have a loop. The Heaviside step function, $\mathcal{H}(\cdot)$, is adopted here because a farther geodesic distance implies a higher expectation of loop closure. Here, $\tau(\cdot)$ denotes a reciprocal of a linear function because the distance between nodes increases along with the $v_{\max}$. Because the yawing considerably harms the pose estimation performance, especially VIO, it is included as a penalizing factor. Accordingly, the set of nodes that are likely to be a loop with high likelihood is defined as follows:

$$\mathcal{N}_K^{LC} = \{n_{k'} \in \mathcal{N}_{K-1} \mid \|\vec{p}_{n_{k'}} - \vec{p}_{n_K}\| \leq r_{\max}, \\ f(\mathcal{X}|\Theta_K) > \kappa(v_{\max})\} \quad (8)$$

where $\kappa(\cdot)$ denotes a linear function to compensate for the aforementioned potential risks of yawing. To maximize the likelihood of the loop, the optimal node $n_{LC}^*$ is selected from $\mathcal{N}_K^{LC}$, if not empty, as follows:

$$n_{LC}^* = \underset{n_{k''} \in \mathcal{N}_K^{LC}}{\operatorname{argmax}} (f(\mathcal{X}|\Theta_K)). \quad (9)$$

The path $\xi_{LC}$ from $n_K$ to $n_{LC}^*$ is obtained from A* search again and refined into the minimum snap trajectory before tracking. When tracking $\xi_{LC}$, the yaw of the UAV is controlled so that it matches the one saved in $n_{LC}^*$ to increase the chance of loop-closing. The explained likelihood-based selection for active loop-closing is depicted in Fig. 4.

### H. Trajectory Refinement

As previously explained, the best global path $\xi_G$ and active loop-closing path $\xi_{LC}$ are merely the sets of edges; however, the path consists of multiple straight lines that are not functions of time, i.e., it is not a trajectory. Therefore, these paths are refined into a minimum snap trajectory before tracking. To solve the coefficients of each trajectory, the desired conditions are set as follows:

$$\begin{cases} v_{\max_i} = v_{\max} \\ T_i = \frac{\ell_{n_i, n_{i+1}}}{v_{\max} T} T \\ \vec{p}(t_{0i}) = \vec{p}_{n_i} \\ \vec{p}(t_{0i} + T_i) = \vec{p}_{n_{i+1}}, \quad i = 0, 1, \ldots, Q \end{cases} \quad (10)$$

where $\vec{p}(t_{0i})$ and $\vec{p}(t_{0i} + T_i)$ denote the initial and the final position of the trajectory to be obtained, respectively. $Q$

denotes the number of elements in the set of edges, $\xi_G$ or $\xi_{LC}$. The velocity is maintained to match $v_{\max}$, which is used in the local exploration. The interval $T_i$ is set in proportion to the length of the trajectory. Given the constraints, the coefficients of the polynomial can be calculated in the same way with the Peacock Trajectory. For better understanding, the best global paths before and after *Trajectory Refinement* are visualized in Fig. 3(b).

## IV. EXPERIMENTS AND RESULTS

### A. Experimental Setting

To evaluate the performance of REAL, simulation-based experiments and a real-world experiment are conducted. For simulation environments, a small-scale maze map and the large-scale mine map[1] based on the Gazebo simulator [21] are used to check whether algorithms are applicable to various scales, which are shown in Fig. 5. For pair comparison, the experiments are repeated ten times in each environment for each $v_{\max}$.

REAL with and without active loop-closing are denoted as "Ours w/ LC" and "Ours w/o LC". REAL is quantitatively compared with the state-of-the-art methods; namely, NBVP [5], GBP [6], MBP [7], and AEP [9]. For the pose estimation, VINS-Stereo [16] is adopted due to its higher accuracy, which was analyzed in [17].

The experimental parameters are represented in Table I.

TABLE I
THE EXPERIMENTAL PARAMETERS

| Parameter | Value | Parameter | Value |
|---|---|---|---|
| FoV | 90×67.5 [°] | $\varrho_{\text{oct}}$ | 0.3×0.3×0.3 [m] |
| $(r_{\max}, \ell_{\text{traj}})_{\text{small}}$ | (6.0, 2.5) [m] | $v_{\max}$ | 0.75, 1.5, 2.5, 3.5 [m/s] |
| $(r_{\max}, \ell_{\text{traj}})_{\text{large}}$ | (8.0, 3.5) [m] | $\dot{\psi}_{\max}$ | 1.0 [rad/s] |

### B. Error Metrics

To evaluate our proposed method quantitatively, total exploration time, $T_{\exp}$, and Root Mean Square Error (RMSE), $\delta_{\text{pose}}$ are introduced, as well as each parameter whose mean and standard deviation for repeated experiments are measured. To place more emphasis on the success of exploration, if no progress occurred for 5 minutes during exploration, then it was determined to be stuck. Note that if stuck or crash happens, we set $T_{\exp} = \infty$ and put the number of failures into the bracket. Likewise, if pose estimation diverges, $\delta_{\text{pose}}$ is set to $\infty$ and the number of divergences is put into the bracket.

### C. Effect of Active Loop-Closing

Active loop-closing resulted in slower exploration at $v_{\max} = 0.75$ m/s on a small-scale map and at $v_{\max} = 2.5$ m/s on a large-scale map, as can be seen in Fig. 6 and Table II. However, except for those cases, "Ours w/ LC" showed faster speed and higher pose estimation accuracy. Because active loop-closing connected more nodes so that the more efficient, shorter paths could be extracted, whereas redundant revisiting is strictly prevented in the local and global exploration planner. Therefore, it can be said that active loop-closing increases not only the accuracy of the pose estimation but the speed of exploration as well.

[1]https://github.com/engcang/gazebo_maps

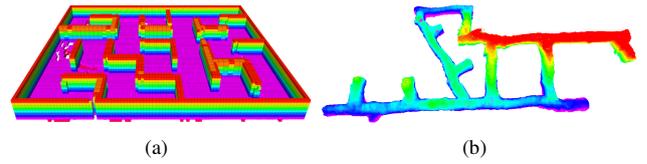

Fig. 5. The OctoMap of the (a) small- and (b) large-scale simulation environments with 28×24×3 and 150×70×12 m³ dimensions, respectively. The colors used in the maps represent the corresponding heights.

### D. Comparison with State-of-the-Art Methods

**Small-scale Environment** NBVP and MBP showed many unnecessary movements due to their randomness, which occasioned the accumulation of the drift of localization (see Fig. 6 and Fig. 7). The aggressive motion primitives in MBP even resulted in the divergences of the localization at the slowest $v_{\max}$ and crashes at the fastest $v_{\max}$, despite having tuned many parameters. GBP was the second-fastest among comparative algorithms thanks to its graph-based global planner; however, it showed divergences of the pose estimation due to its infeasible trajectory planning without consideration of the kinodynamics of the UAV, which showed a worse result than the simple straight lined-paths from NBVP. By leveraging the advantages of the sampling-based and the frontier-based methods, AEP showed the best exploration speed and localization accuracy among comparative algorithms; however, it showed the UAV hovered idly for a few seconds while computing the complex local and global planning, which was one of the main reasons for the slower speed than ours. Thanks to efficient graph-based global planning, high velocity from Peacock Trajectory, and active loop-closing, the proposed approach outperformed state-of-the-art algorithms both in the speed and the accuracy of the localization.

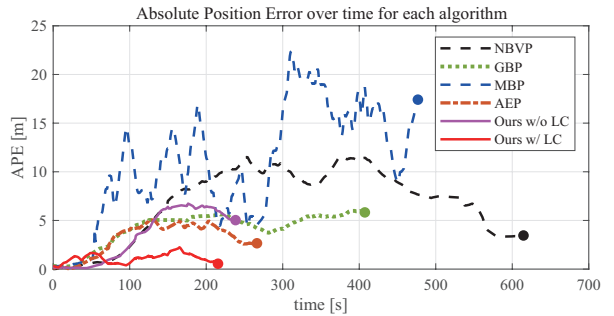

Fig. 6. The Absolute Pose Error (APE) over exploration time for each algorithm in the small-scale simulation environment at $v_{\max} = 0.75$ m/s. The lower APEs and the shorter durations are better. The plotted results are the median cases of ten runs. The circles represent the end time.

**Large-scale Environment** For the large-scale map, none of the comparative groups finished the whole ten runs but only the proposed approach. Only GBP completed a few of ten runs thanks to its graph-based global planner; however, every gain computation took around ten seconds and the UAV eventually become stuck for most of ten runs. AEP stopped early in the large-scale map because computational time took longer than its own upper limit and the algorithm judged that the exploration had finished. MBP and NBVP showed the same unnecessary back-and-forth behaviors, so they were stuck for all of ten runs. The proposed algorithm, REAL, easily completed the large-scale map thanks to its trajectory, which assumes the maximum velocities at the initial and final position, and graph-based exploration that is conducted only for redundant visiting situations. In addition, by actively closing the loop, it showed reasonable pose estimation performance, sufficient to be applied in real-world applications.

TABLE II
THE RESULTS OF SIMULATION EXPERIMENTS (THE SHADED AREAS REPRESENT THE BEST RESULTS.)

| Map | $v_{max}$ | NBVP [5] $T_{exp}$ | $\delta_{pose}$ | GBP [6] $T_{exp}$ | $\delta_{pose}$ | MBP [7] $T_{exp}$ | $\delta_{pose}$ | AEP [9] $T_{exp}$ | $\delta_{pose}$ | Ours w/o LC $T_{exp}$ | $\delta_{pose}$ | Ours w/ LC $T_{exp}$ | $\delta_{pose}$ |
|---|---|---|---|---|---|---|---|---|---|---|---|---|---|
| small | 0.75m/s | 604±89 | 7.27±3.07 | 411±47 | 3.01±0.78 | 543±141 | ∞ (3) | 288±28 | 2.54±1.08 | 238±20 | 4.63±1.46 | 249±21 | 1.35±0.29 |
| small | 1.5m/s | 562±131 | 5.58±2.03 | 244±31 | ∞ (3) | 440±104 | ∞ (10) | 220±33 | 3.24±1.65 | 173±12 | 3.36±1.39 | 160±16 | 2.09±1.56 |
| small | 2.5m/s | 512±117 | 8.98±3.58 | 207±19 | ∞ (8) | ∞ (7) | × | 202±33 | 3.77±2.35 | 149±25 | 3.62±1.41 | 142±9 | 1.65±0.51 |
| large | 2.5m/s | ∞ (10) | × | ∞ (3) | × | ∞ (10) | × | ∞ (10) | × | 346±18 | 9.70±2.31 | 357±23 | 2.85±1.29 |
| large | 3.5m/s | ∞ (10) | × | ∞ (5) | × | ∞ (10) | × | ∞ (10) | × | 301±11 | 5.81±3.17 | 299±36 | 4.49±2.22 |

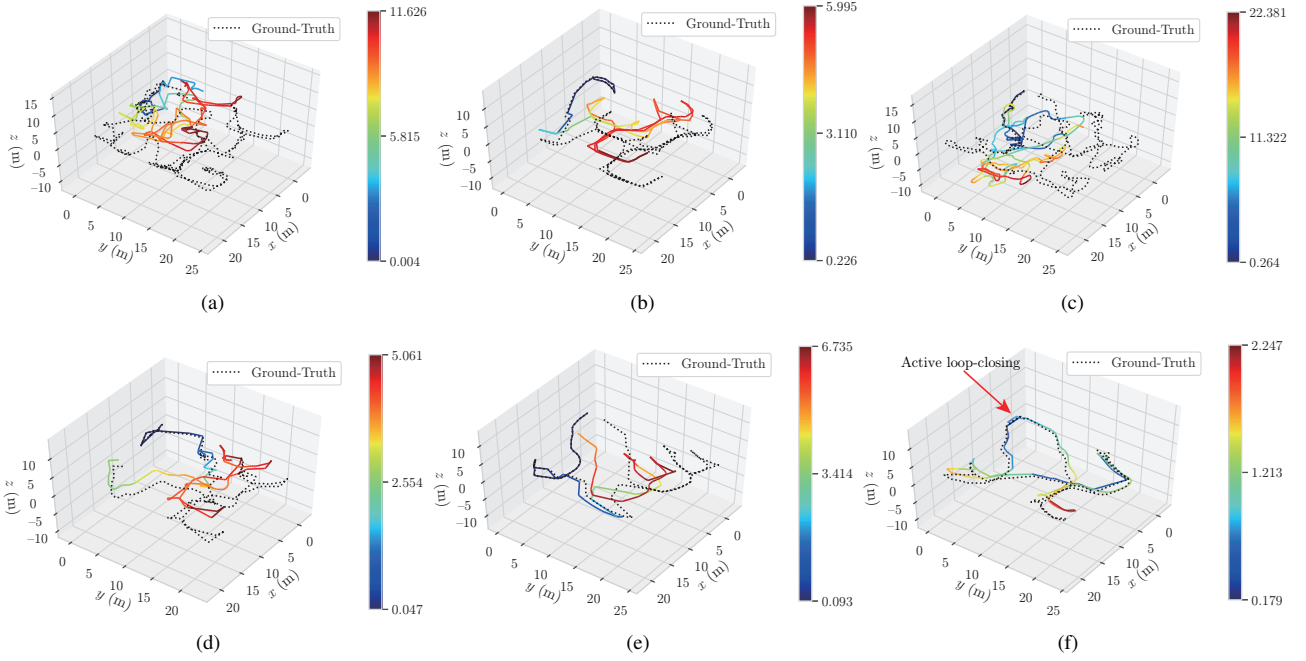

Fig. 7. (a) NBVP; (b) GBP; (c) MBP; (d) AEP; (e) Ours w/o LC; (f) Ours w/ LC. The estimated flight trajectories from VINS-Stereo [16] (colored) versus ground truth (dashed black line) in a small-scale environment at $v_{max}$ = 0.75 m/s. The result shown is the median case of ten runs, which is plotted with EVO [25]. Unnecessary movements can be easily checked at NBVP and MBP from flight trajectories (ground truth). The loop closure can be found at Ours w/ LC thanks to its active loop-closing.

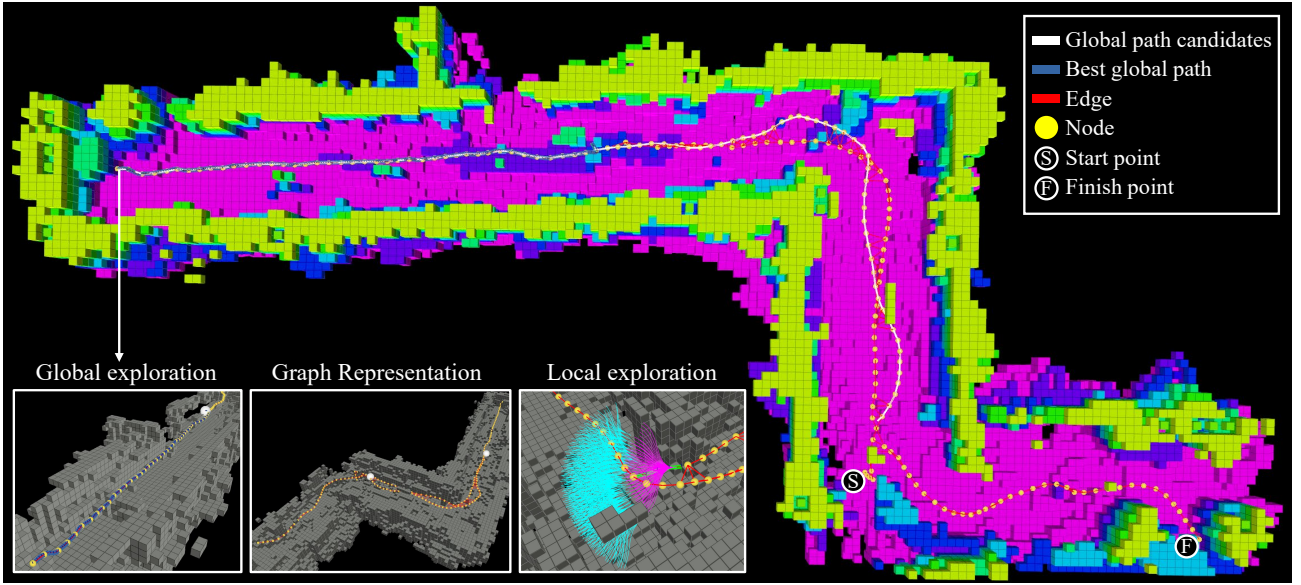

Fig. 8. The result of the exploration in a real-world environment. The occupancy grid map, which is truncated at 2.2 m for better visualization, is represented with the colored map corresponding to its height. The UAV explored the environment at $v_{max}$ = 0.5m/s. At the dead-end in the left part of the figure, the global exploration planning was conducted using the graph and the frontier clusters.

*E. Validation in Real-World Environment*

To validate the capability of REAL, a large-scale real-world experiment was conducted. The UAV equipped with an RGB-D camera (Intel D435i) explored a 38×19×3 m$^3$ bounded, underground environment. The $v_{\max}$ and $\dot{\psi}_{\max}$ were set to 0.5 m/s and 1.0 rad/s, respectively. An Intel T265 camera was used for the pose estimation. Data processing and the exploration algorithm were run onboard. The instance and the result of the experiment can be seen in Fig. 1 and Fig. 8. The UAV autonomously mapped the environment completely. The video of the experiment can be viewed at https://youtube.com/playlist?list=PLvgPHeVm_WqJCRyua8g4d5cw8yUaRlzNL.

## V. CONCLUSIONS

The current paper proposed utilizing a UAV's kinodynamics-based trajectory and a graph-based global exploration planner to maximize the efficiency of spatial visiting and hence the speed of exploration. At the same time, the drift of the pose estimation was minimized thanks to active loop-closing based on the likelihood. Compared to the state-of-the-art algorithms, the proposed approach outperforms in both of the speed and pose estimation accuracy during the flight, which were demonstrated in small- and large-scale simulation maps. The capability of the algorithm was validated in a real-world experiment. Future works will extend to the different types of trajectories to be applied in diverse robots (e.g., legged robots).